\pdfoutput=1

\documentclass[11pt]{article}

\usepackage{acl}

\usepackage{times}
\usepackage{latexsym}

\usepackage[T1]{fontenc}

\usepackage[utf8]{inputenc}

\usepackage{microtype}

\usepackage{amsmath}
\usepackage{amsfonts}
\usepackage{booktabs, multirow}

\usepackage{caption}
\usepackage{subcaption}
\usepackage{graphicx}

\usepackage{comment}
\DeclareMathOperator*{\argmax}{arg\,max}

\usepackage{etoolbox}
\patchcmd{\quote}{\rightmargin}{\leftmargin 1em \rightmargin}{}{}

\usepackage{enumitem}
\setlist[itemize]{%
itemindent=0pt,%
leftmargin=*,%
listparindent=-\leftmargin%
}

\newcommand\blfootnote[1]{%
  \begingroup
  \renewcommand\thefootnote{}\footnote{#1}%
  \addtocounter{footnote}{-1}%
  \endgroup
}

\title{Quality-Aware Decoding for Neural Machine Translation}

\author{
\textbf{Patrick Fernandes}$^{*, 1,2,3}$  \quad
\textbf{António Farinhas}$^{*, 2,3}$  \quad
\textbf{Ricardo Rei}$^{2,4,5}$  \quad
\textbf{José G. C. de Souza}$^{5}$ \\
\textbf{Perez Ogayo}$^{1}$ \quad
\textbf{Graham Neubig}$^{1}$ \quad
\textbf{André F. T. Martins}$^{2,3,5}$ \quad
\\
$^1$Carnegie Mellon University\quad
$^2$Instituto Superior Técnico  (Lisbon ELLIS Unit)\quad\\
$^3$Instituto de Telecomunicações\quad
$^4$INESC-ID \quad
$^5$Unbabel\\
{\small \texttt{pfernand@cs.cmu.edu} \phantom{aaaa}\texttt{antonio.farinhas@tecnico.ulisboa.pt}} 
}

\begin{document}
\maketitle
\begin{abstract}
\blfootnote{* Equal contribution.} 
Despite the progress in machine translation quality estimation and evaluation in the last years, decoding in neural machine translation (NMT) is mostly oblivious to this and centers around finding the most probable translation according to the model (MAP decoding), approximated with beam search.  
In this paper, we bring together these two lines of research and propose \emph{quality-aware decoding} for NMT, by leveraging recent breakthroughs in reference-free and reference-based MT evaluation through various inference methods like $N$-best reranking and minimum Bayes risk decoding.
We perform an extensive comparison of various possible {candidate generation} and {ranking} methods across four datasets and two model classes and find that quality-aware decoding consistently outperforms MAP-based decoding  
according  both to state-of-the-art automatic metrics (COMET and BLEURT) and to human assessments. {Our code is available at \url{https://github.com/deep-spin/qaware-decode}.} 

\end{abstract}

\section{Introduction}
\label{sec:introduction}


The most common procedure in neural machine translation (NMT) is to train models using maximum likelihood estimation (MLE) at training time, and to decode with beam search at test time, as a way to approximate maximum-a-posteriori (MAP) decoding.
However, several works have questioned the utility of model likelihood as a good proxy for translation quality \citep{koehn-knowles-2017-six,ott2018analyzing, stahlberg-byrne-2019-nmt, eikema-aziz-2020-map}.  
In parallel, significant progress has been made in methods for quality estimation and evaluation of generated translations \cite{specia-etal-2020-findings-wmt, mathur-etal-2020-results}, but this progress is, by and large, not yet reflected in either training or decoding methods.
Exceptions such as minimum risk training \citep{shen-etal-2016-minimum, edunov-etal-2018-classical} come at a cost of more expensive and unstable training, often with modest quality improvements.  

\begin{figure}[t]
     \centering
      \includegraphics[width=\columnwidth]{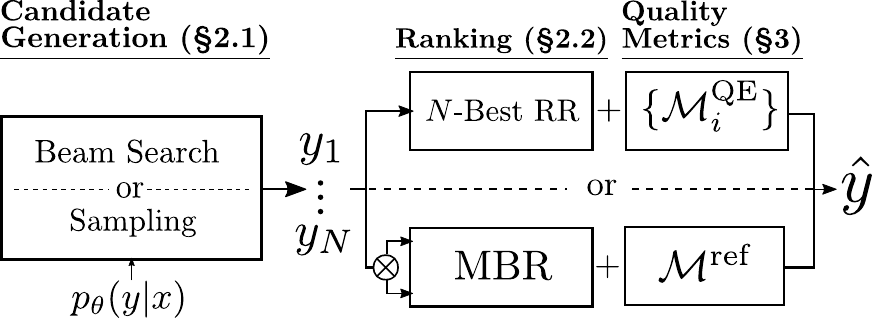}
     \caption{\label{fig:main-fig} 
     Quality-aware decoding framework. First, translation candidates are \textit{generated} according to the model. Then, using  reference-free and/or reference-based MT metrics, these candidates are \textit{ranked}, and the highest ranked one is picked as the final translation.}
     \vspace{-0.4em}
\end{figure}

An appealing alternative is to modify the decoding procedure only, separating it into two stages: \textit{candidate generation} (\S\ref{subsec:candidate-generation}; where candidates are generated with beam search or sampled from the whole distribution) and \textit{ranking} (\S\ref{subsec:ranking}; where they are scored using a quality metric of interest, and the translation with the highest score is picked). 
This strategy has been explored in approaches using $N$-best reranking \citep{ng-etal-2019-facebook,bhattacharyya-etal-2021-energy} and minimum Bayes risk (MBR) decoding \citep{shu2017laterstage,eikema2021samplingbased, muller2021understanding}. 
While this previous work has exhibited promising results, it has mostly focused on optimizing lexical metrics such as BLEU or METEOR \citep{papineni-etal-2002-bleu,lavie2009meteor}, which have  limited correlation with human judgments \cite{mathur-etal-2020-tangled, freitag-et-al-2020-experts}. 
Moreover, a rigorous apples-to-apples comparison among this suite of  techniques and their variants is still missing, even though they share similar building blocks.

Our work fills these gaps by asking the question:
\begin{quote}
``\emph{Can we leverage recent advances in MT quality evaluation to generate better translations? If so, how can we most effectively do so?}''
\end{quote}
To answer this question, we systematically explore NMT decoding using a suite of ranking procedures.  We take advantage of recent state-of-the-art learnable metrics, both reference-based, such as COMET and BLEURT \cite{rei-etal-2020-comet, sellam-etal-2020-bleurt}, and reference-free (also known as \textit{quality estimation}; QE), such as TransQuest and OpenKiwi \cite{ranasinghe-etal-2020-transquest,kepler-etal-2019-openkiwi}. We compare different ranking strategies under a unified  framework, which we name \textbf{quality-aware decoding} (\S\ref{sec:quality-aware}). 
First, we analyze the performance of decoding using $N$-best reranking, both \textit{fixed} according to a single metric and \textit{learned} using multiple metrics, where the coefficients for each metric are optimized according to a reference-based metric. Second, we explore ranking using reference-based metrics directly through MBR decoding. 
Finally, to circumvent the expensive computational cost of the latter when the number of candidates is large, we develop a two-stage ranking procedure, where we use $N$-best reranking to pick a subset of the candidates to be ranked through MBR decoding. 
We explore the interaction of these different ranking methods with various candidate generation procedures including beam search, vanilla sampling, and nucleus sampling.

Experiments with two model sizes and four datasets (\S\ref{sec:experiments}) reveal that while MAP-based decoding appears competitive when evaluating with lexical-based metrics (BLEU and ChrF), the story is very different with state-of-the-art evaluation metrics, where quality-aware decoding shows significant gains, both with $N$-best reranking and MBR decoding. 
We perform a human-study to more faithfully evaluate our systems and find that, while performance on learnable metrics is not always predictive of the best system, quality-aware decoding usually results in translations with higher quality than MAP-based decoding.

\section{Candidate Generation and Ranking}

We start by reviewing some of the most commonly used methods for both candidate generation and ranking under a common lens.

\subsection{Candidate Generation}
\label{subsec:candidate-generation}
An NMT model defines a probability distribution $p_\theta(y|x)$ over a set of hypotheses $\mathcal{Y}$, conditioned on a source sentence $x$, where $\theta$ are learned parameters.
A translation is typically predicted using MAP decoding, formalized as
\begin{equation}
    \hat{y}_{\textsc{map}} = \argmax_{y\in\mathcal{Y}} ~ \log p_\theta(y|x).
\end{equation}
In words, MAP decoding searches for the most probable translation under $p_\theta(y|x)$, \textit{i.e.}, the mode of the model distribution. Finding the exact $\hat{y}_{\mathrm{MAP}}$  is intractable since the search space $\mathcal{Y}$ is combinatorially large, thus, approximations like \textbf{beam search} \citep{graves2012sequence, sutskever2014sequence} are used. 
However, it has been shown that the translation quality \emph{degrades} for large values of the beam size 
\citep{koehn-knowles-2017-six, yang-etal-2018-breaking, murray-chiang-2018-correcting, meister-etal-2020-beam}, with the empty string often being the true MAP hypothesis \citep{stahlberg-byrne-2019-nmt}.

A stochastic alternative to beam search is to  \emph{draw samples} directly from $p_\theta(y|x)$ with ancestral sampling, optionally with variants that truncate this distribution, such as top-$k$ sampling \citep{fan-etal-2018-hierarchical} 
or \textbf{$p$-nucleus sampling} \citep{Holtzman2020The} -- the latter samples from the smallest set of words whose cumulative probability is larger than a predefined value $p$. 
Deterministic methods combining beam and nucleus search have also been proposed \citep{shaham2021cross}.

Unlike beam search, sampling is not a search algorithm nor a decision rule -- it is not expected for a single sample 
to outperform MAP decoding \citep{eikema-aziz-2020-map}. However, samples from the model can still be useful for alternative decoding methods, as we shall see. 
While beam search focus on high probability candidates, typically similar to each other, sampling  allows for more \emph{exploration}, leading to higher candidate \emph{diversity}.

\subsection{Ranking}
\label{subsec:ranking}
We assume access to a set 
$\mathcal{\bar{Y}} \subseteq \mathcal{Y}$ containing $N$ candidate translations for a source sentence, obtained with one of the generation procedures described in \S\ref{subsec:candidate-generation}.
As long as $N$ is relatively small, it is possible to (re-)rank these candidates in a post-hoc manner, such that the best translation maximizes a given metric of interest.
We highlight two different lines of work 
for ranking in MT decoding: first, \textbf{$N$-best reranking}, using reference-free metrics as features; 
second, \textbf{MBR decoding}, using reference-based metrics. 

\subsubsection{$N$-best Reranking}
\label{subsubsec:N-best}

In its simplest form (which we call \textit{fixed} reranking), a \emph{single} feature $f$ is used  (\textit{e.g.}, an estimated quality score), and the candidate 
that maximizes this score is picked as the final translation, 
\begin{equation}
\label{eq:nbest-reranking-fixed}
    \hat{y}_{\textsc{f-rr}} = \argmax_{y \in \mathcal{\bar{Y}}} ~ f(y).
\end{equation}
When \emph{multiple} features $[f_1,\ldots,f_K]$ are available, one can tune weights  $[w_1,\ldots,w_K]$ for these features 
to maximize a given reference-based evaluation metric on a validation set \citep{och-2003-minimum,duh-kirchhoff-2008-beyond} -- we call this \textit{tuned} reranking.
In this case, the final translation is 
\begin{equation}
\label{eq:nbest-reranking-with-weights}
    \hat{y}_{\textsc{t-rr}} = \argmax_{y \in \mathcal{\bar{Y}}} ~ \textstyle \sum_{k=1}^{K} w_k f_k(y).
\end{equation}


\subsubsection{Minimum Bayes Risk (MBR) Decoding}
\label{subsubsec:MBR}
While the techniques above rely on \textit{reference-free} metrics for the computation of features, 
MBR decoding uses \textit{reference-based} metrics to rank candidates. 
Unlike MAP decoding, 
which searches for the most probable translation, MBR decoding aims to find the translation that maximizes the expected \emph{utility} (equivalently, that minimizes \emph{risk}, \citealt{kumar2002minimum, kumar-byrne-2004-minimum,eikema-aziz-2020-map}).
Let again $\bar{\mathcal{Y}} \subseteq \mathcal{Y}$ be a set containing $N$ hypotheses and $u(y^*, y)$ a utility function measuring the similarity between a hypothesis $y\in\mathcal{Y}$ and a reference $y^* \in \bar{\mathcal{Y}}$ (\textit{e.g}, an automatic evaluation metric such as BLEU or COMET). 
MBR decoding seeks for
\begin{align}\label{eq:MC-expectation}
    \hat{y}_{\textsc{mbr}} &= \argmax_{y\in\bar{\mathcal{Y}}} ~\underbrace{\mathbb{E}_{Y \sim p_\theta(y \mid x)}[ u(Y, y)]}_{\textstyle \approx ~\frac{1}{M}\sum_{j=1}^{M}u(y^{(j)}, y)}, 
\end{align}
where in Eq.~\ref{eq:MC-expectation} the expectation is approximated as a Monte Carlo (MC) sum using model samples $y^{(1)}, \ldots, y^{(M)} \sim p_\theta(y|x)$.%
\footnote{We also consider the case where $y^{(1)}, \ldots, y^{(M)}$ are obtained from nucleus sampling or beam search.
Although the original MC estimate 
is unbiased, these ones are biased.} %
In practice, the translation with the highest expected utility can be computed by comparing each hypothesis $y \in \bar{\mathcal{Y}}$ to all the other hypotheses in the set. 

\section{Quality-Aware Decoding}
\label{sec:quality-aware}

While recent works have explored various combinations of candidate generation and ranking procedures for NMT \cite{lee-etal-2021-discriminative, bhattacharyya-etal-2021-energy, eikema2021samplingbased, muller2021understanding}, they suffer from two limitations:
\begin{itemize}
    \item The ranking procedure is usually based on simple lexical-based metrics (BLEU, chrF,  METEOR). Although these metrics 
    are well established and 
    inexpensive to compute, they correlate poorly with human judgments at segment level \cite{mathur-etal-2020-results, Freitag2021}. 
    \item  Each work independently explores $N$-best reranking or MBR decoding, making unclear which method produces better translations.
\end{itemize}

In this work, we hypothesize that using more powerful metrics in the ranking procedure may lead to better quality translations.
We propose a unified framework for ranking with both reference-based  (\S\ref{reference-based-metrics}) and reference-free metrics (\S\ref{subsec:Reference-free Metrics}), independently of the candidate generation procedure.
We explore four methods with different computational costs for a given number of candidates, $N$.

\paragraph{Fixed $N$-best Reranker.}
An $N$-best reranker using a single reference-free metric (\S\ref{subsec:Reference-free Metrics}) as a feature, according to Eq.~\ref{eq:nbest-reranking-fixed}. The computational cost of this ranker is $\mathcal{O}(N\times C_{\mathcal{M}^{\mathrm{QE}}})$, where $C_{\mathcal{M}^{\mathrm{QE}}}$ denotes the cost of running an evaluation with a metric $\mathcal{M}^\mathrm{QE}$. 

\paragraph{Tuned $N$-best Reranker.}
An $N$-best reranker using as features \emph{all} the reference-free metrics in \S\ref{subsec:Reference-free Metrics}, along with the model log-likelihood $\log p_\theta(y | x)$. 
The weights in Eq.~\ref{eq:nbest-reranking-with-weights} are optimized to maximize a given reference-based metric $\mathcal{M}^{\mathrm{ref}}$
using MERT \citep{och-2003-minimum}, a coordinate-ascent optimization algorithm widely used in previous work. Note that $\mathcal{M}^\mathrm{ref}$ is used for tuning only; at test time, only reference-free metrics are used. Therefore, 
the decoding cost is $\mathcal{O}(N\times \sum_i C_{\mathcal{M}^{\mathrm{QE}}_i})$. 

\paragraph{MBR Decoding.} Choosing as the utility function a reference-based metric $\mathcal{M}^{\mathrm{ref}}$ (\S\ref{reference-based-metrics}), we estimate the utility using a simple Monte Carlo sum, as shown in Eq.~\ref{eq:MC-expectation}. 
The estimation requires computing pairwise comparisons and thus the cost of running MBR decoding is $\mathcal{O}(N^2\times C_{\mathcal{M}^{\mathrm{ref}}})$. 
    
\paragraph{$N$-best Reranker $\rightarrow$ MBR.} 
Using a large number of samples in MBR decoding is  expensive due to its quadratic cost.
To circumvent this issue, we explore a \emph{two-stage} ranking approach: we first rank all the candidates using a tuned $N$-best reranker, followed by MBR decoding using the top $M$ candidates. The computational cost becomes $\mathcal{O}(N\times \sum_i C_{\mathcal{M}_i} + M^2\times C_{\mathcal{M}^{\mathrm{ref}}})$. The first ranking stage \emph{prunes} the candidate list to a smaller, higher quality subset, making possible a more accurate estimation of the utility with less samples, and potentially allowing a better ranker than \emph{plain} MBR for almost the same computational budget.

\subsection{Reference-based Metrics}
\label{reference-based-metrics}

Reference-based metrics are the standard way to evaluate MT systems; the most used ones rely on the lexical overlap between hypotheses and reference translations  \cite{papineni-etal-2002-bleu,lavie2009meteor,popovic-2015-chrf}.
However, lexical-based approaches have important limitations:
they have difficulties recognizing correct translations that are paraphrases of the reference(s); they ignore the source sentence, an important indicator of meaning for the translation; and they do not always correlate well with human judgments, particularly at segment-level~\cite{Freitag2021}. 


In this work, apart from BLEU (computed using SacreBLEU\footnote{ \texttt{nrefs:1|case:mixed|eff:no|tok:13a\\|smooth:exp|version:2.0.0}} \cite{post-2018-call}) and chrF, we use the following state-of-the-art trainable reference-based metrics for both ranking and performance evaluation of MT systems:

\begin{itemize}
    \item BLEURT \citep{sellam-etal-2020-bleurt, pu-etal-2021-learning}, trained to regress on human direct assessments (DA; \citealt{graham-etal-2013-continuous}). 
    We use the largest multilingual version, \textit{BLEURT-20}, based on the RemBERT model \cite{Chung2021RethinkingEC}.
    \item COMET \citep{rei-etal-2020-comet}, based on XLM-R \cite{conneau-etal-2020-unsupervised}, trained to regress on quality assessments such as DA using both the reference and the source to assess the quality of a given translation. We use the publicly available model developed for the WMT20 metrics shared task (\textit{wmt20-comet-da}).
\end{itemize}

These metrics have shown much better correlation at segment-level than previous lexical metrics in WMT metrics shared tasks~\cite{mathur-etal-2020-results, Freitag2021}. Hence, as discussed in \S\ref{subsec:ranking}, they are good candidates to be used either \emph{indirectly} as an optimization objective for learning the tuned reranker's feature weights, or \emph{directly} as a utility function 
in MBR decoding. In the former, the higher the metric correlation with human judgment, the better the translation picked by the tuned reranker. In the latter, we approximate the expected utility in Eq.~\ref{eq:MC-expectation} by letting a candidate generated by the model be a reference translation -- a suitable premise \emph{if} the model is good in expectation.




\subsection{Reference-free Metrics}
\label{subsec:Reference-free Metrics}

MT evaluation metrics have also been developed for the case where references are not available -- they are called 
\textit{reference-free} or \textit{quality estimation} (QE) metrics. 
In the last years, considerable improvements have been made to such metrics, with state-of-the-art models having increasing correlations with human annotators \citep{Freitag2021, Specia2021}. 
These improvements enable the use of such models for ranking translation hypotheses in a more reliable way than before. 


In this work, we explore four recently proposed  reference-free metrics as features for $N$-best reranking, 
all at the sentence-level:

\begin{itemize}
    \item COMET-QE \citep{rei-etal-2020-unbabels}, a reference-free version of COMET (\S\ref{reference-based-metrics}). It was the winning submission for the QE-as-a-metric subtask of the WMT20 shared task \citep{mathur-etal-2020-results}.
    \item TransQuest \citep{ranasinghe-etal-2020-transquest}, the winning submission for the sentence-level DA prediction subtask of the WMT20 QE shared task \citep{specia-etal-2020-findings-wmt}. Similarly to COMET-QE this metric predicts a DA score.
    \item MBART-QE \citep{Zerva-etal-2021-ist}, based on the mBART \cite{liu-etal-2020-multilingual-denoising} model, trained to predict both the \textit{mean} and the \textit{variance} of DA scores. It was a top performer in the WMT21 QE shared task \citep{Specia2021}.
    \item OpenKiwi-MQM \citep{kepler-etal-2019-openkiwi, rei-etal-2021-references},  based on XLM-R, trained to predict the \emph{multidimensional quality metric} (MQM; \citealt{mqm}).\footnote{MQM annotations are expert-level type of annotations more fine-grained then DA, with individual errors annotated.}
    This reference-free metric was ranked second on the QE-as-a-metric subtask from the WMT 2021 metrics shared task.
\end{itemize}

\section{Experiments}\label{sec:experiments}

\begin{figure*}[t]
     \centering
     \begin{subfigure}[b]{0.5\textwidth}
         \centering
         \includegraphics[width=\textwidth]{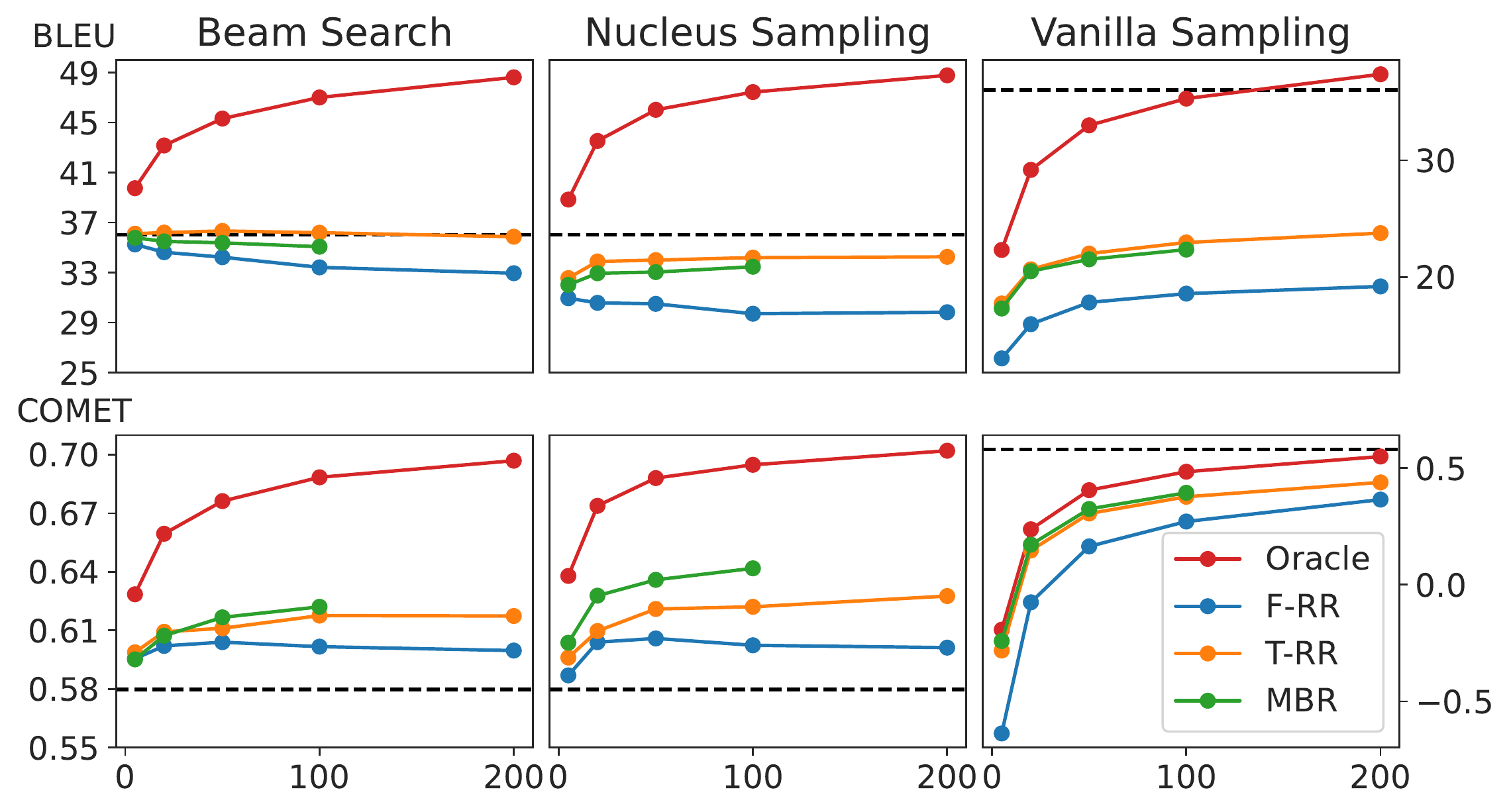}
     \end{subfigure}
     \hspace{-0.3cm}
     \begin{subfigure}[b]{0.5\textwidth}
         \centering
         \includegraphics[width=\textwidth]{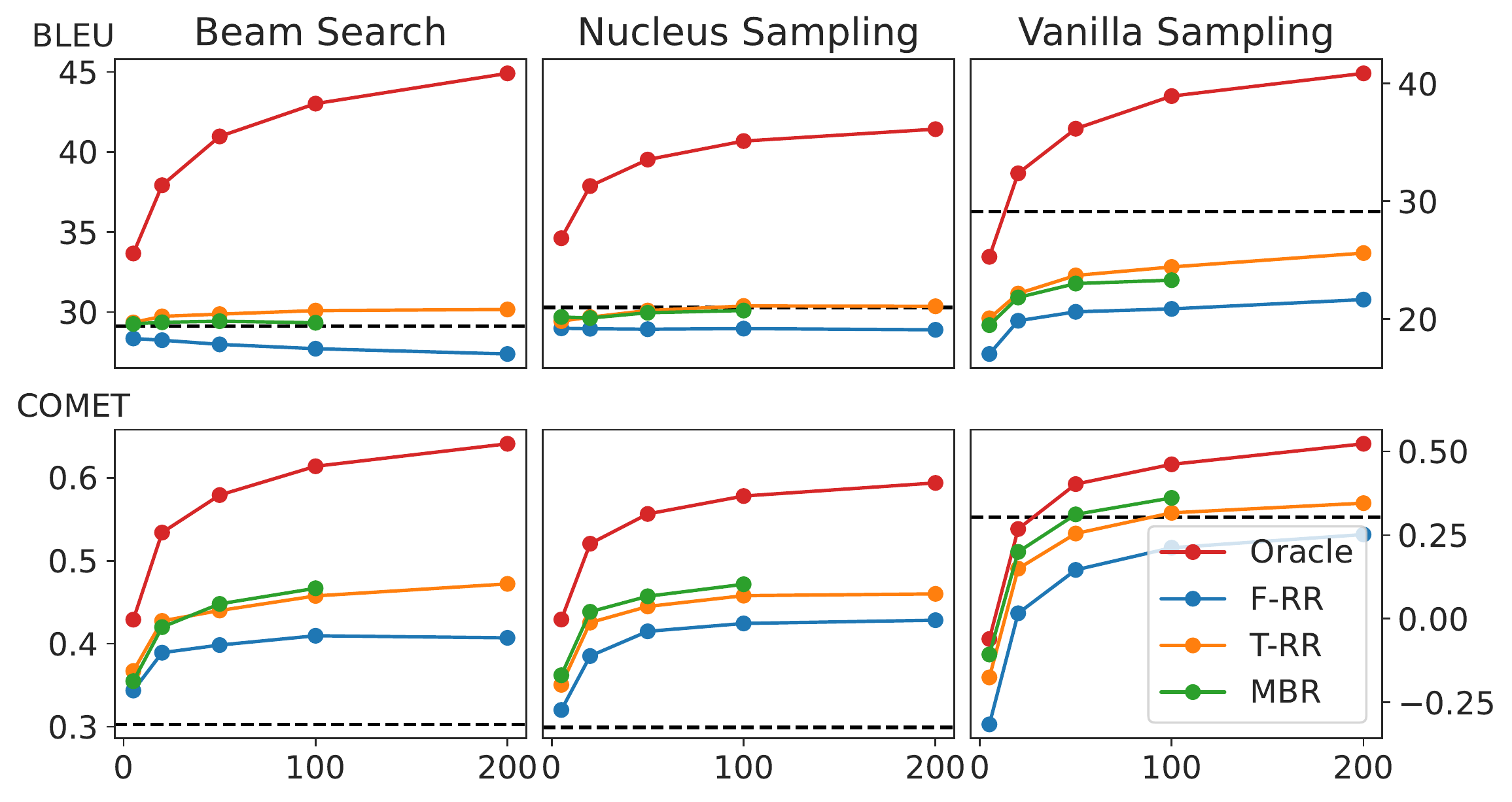}
     \end{subfigure}
     \caption{\label{fig:candidates-analysis} 
     Values for BLEU (top) and COMET (bottom) for $\text{EN}\rightarrow \text{DE}$ as we increase the number of candidates for different generation and ranking procedures, as well as oracles with the respective metrics, for the \emph{large} (left) and \emph{small} (right) models. Baseline values (with beam size of 5) are marked with a dashed horizontal line. } 
\end{figure*}

\subsection{Setup}

We study the benefits of
quality-aware decoding 
over MAP-based decoding in two  regimes:
\begin{itemize}
    \item A high-resource, unconstrained, setting with \emph{large} transformer models (6 layers, 16 attention heads, 1024 embedding dimensions, and 8192 hidden dimensions) trained by \citet{ng-etal-2019-facebook} for the WMT19 news translation task \citep{barrault-etal-2019-findings}, using English to German  ($\text{EN}\rightarrow \text{DE}$) and  English to Russian ($\text{EN}\rightarrow \text{RU}$) language pairs. These models were trained on over 20 million parallel and 100 million back-translated sentences, being the winning submissions of that year's shared task. We consider the non-ensembled version of the model and use  \textit{newstest19} for validation and \textit{newstest20} for  testing.
    \item A more constrained scenario with a \emph{small} transformer model (6 layers, 4 attention heads, 512 embedding dimensions, and 1024 hidden dimensions) trained from scratch in \textit{Fairseq}  \citep{ott-etal-2019-fairseq} on the smaller IWSLT17 datasets \cite{cettolo-2012} for English to German ($\text{EN}\rightarrow \text{DE}$) and English to French ($\text{EN}\rightarrow \text{FR}$), each with a little over 200k training examples. We chose these datasets because they have been extensively used in previous work \cite{bhattacharyya-etal-2021-energy} and smaller model allows us to answer questions about how the training methodology affects ranking performance (see \S~\ref{subsubsec:label-smoothing}). Further training details can be found in Appendix~\ref{sec:training_details}.

\end{itemize}

We use beam search with a beam size of 5 as our decoding baseline because we found that it resulted in better or similar translations than larger beam sizes.
For tuned \textit{N}-best reranking, we use Travatar's \cite{neubig13travatar} implementation of MERT \citep{och-2003-minimum} to optimize the weight of each feature, as described in \S\ref{subsec:Reference-free Metrics}.
Finally, we evaluate each system using the metrics discussed in \S\ref{reference-based-metrics}, along with BLEU and chrF \citep{popovic-2015-chrf}.

\subsection{Results}

Overall, given all the metrics, candidate generation, and ranking procedures, we evaluate over 150 systems per dataset. We report subsets of this data separately to answer specific research questions, and defer to Appendix \ref{sec:appendix_full_results} for additional results.

\subsubsection{Impact of Candidate Generation}

First, we explore the impact of the candidate generation procedure and the number of candidates. 

\paragraph{\textit{Which candidate generation method works best, beam search or sampling?}}

We generate candidates with {beam search}, {vanilla sampling}, and {nucleus sampling}. 
For the latter, we use $p=0.6$ based on early results showing improved performance for all metrics.\footnote{We picked nucleus sampling over top-$k$ sampling because it allows varying support size and has outperformed top-$k$ in text generation tasks \citep{Holtzman2020The}.} %
For $N$-best reranking, we use up to 200 samples; for MBR decoding, due to the quadratic computational cost, we use up to 100.

Figure~\ref{fig:candidates-analysis} shows BLEU and COMET for different candidate generation and ranking methods for the $\text{EN}\rightarrow \text{DE}$ WMT20 and IWSLT17 datasets, with increasing number of candidates. The baseline is represented by the dashed line. To assess the performance \textit{ceiling} of the rankers, we also report results with an \textit{oracle} ranker for the reported metrics, picking the candidate that maximizes it. For the \textit{fixed} $N$-best reranker, we use COMET-QE as a metric, albeit the results for other reference-free metrics are similar. Performance seems to scale well with the number of candidates, particularly for vanilla sampling and for the \emph{tuned} $N$-best reranker and MBR decoder. \citep{lee-etal-2021-discriminative, muller2021understanding}. However, all the rankers using vanilla sampling severely under-perform the baseline in most cases (see also \S\ref{subsubsec:label-smoothing}). In contrast, the rankers using beam search or nucleus sampling are competitive or outperform the baseline in terms of BLEU, and greatly outperform it in terms of COMET.
For the larger models, we see that the performance according to the lexical metrics degrades with more candidates. In this scenario, rankers using nucleus sampling seem to have an edge over the ones that use beam search for COMET.

Based on the findings above, and due to generally better performance of COMET over BLEU for MT evaluation \citep{kocmi2021ship}, in following experiments we use nucleus sampling with the \emph{large} model and beam search with the \emph{small} model. 

\begin{table*}[t]\centering
\small
\begin{tabular}{lrrrrrrrrr}\toprule
&\multicolumn{4}{c}{Large (WMT20)} &\multicolumn{4}{c}{Small (IWSLT)} \\\cmidrule(lr){2-5} \cmidrule(lr){6-9}
&BLEU &chrF &BLEURT &COMET &BLEU &chrF &BLEURT &COMET \\\midrule
Baseline &\textbf{36.01} &63.88 &0.7376 &0.5795 &29.12 &56.23 &0.6635 &0.3028  \\
\midrule
F-RR w/ COMET-QE &29.83 &59.91 &\underline{0.7457} &\underline{0.6012} &\underline{27.38} &54.89 &\underline{0.6848} &\underline{0.4071}  \\
F-RR w/ MBART-QE & \underline{32.92} &\underline{62.71} &{0.7384} &0.5831 &{27.30} &\underline{55.62} &0.6765 &0.3533 \\
F-RR w/ OpenKiwi &{30.38} &59.56 &0.7401 &0.5623 &25.35 &51.53 &0.6524 &0.2200  \\
F-RR w/ Transquest &{31.28} &{60.94} &0.7368 &0.5739&26.90 &54.46 &0.6613 &0.2999  \\
\midrule
T-RR w/ BLEU &\underline{35.34} &\underline{63.82} &0.7407 &0.5891 &\textbf{\underline{30.51}} &\textbf{\underline{57.73}} &0.7077 &0.4536  \\
T-RR w/ BLEURT &33.39 &62.56 &\underline{0.7552} &0.6217 &30.16 &57.40 &\underline{0.7127} &\underline{0.4741} \\
T-RR w/ COMET &34.26 &63.31 &0.7546 &\underline{0.6276} &30.16 &57.32 &0.7124 &0.4721  \\
\midrule
MBR w/ BLEU & \underline{34.94} &\underline{63.21} &0.7333 &0.5680 &29.25 &56.36 &0.6619 &0.3017  \\
MBR w/ BLEURT  & 32.90 &62.34 &\underline{0.7649} &0.6047 &28.69 &56.28 &\underline{0.7051} &0.3799 \\
MBR w/ COMET &33.04 &62.65 &0.7477 &\underline{0.6359} &\underline{29.43} &\underline{56.74} &0.6882 &\underline{0.4480}  \\
\midrule
T-RR+MBR w/ BLEU & \underline{35.84} &\textbf{\underline{63.96}} &0.7395 &0.5888 &\underline{30.23} &\underline{57.34} &0.6913 &0.3969  \\
T-RR+MBR w/ BLEURT & 33.61 &62.95 &\textbf{\underline{0.7658}} &0.6165 &29.28 &56.77 &\textbf{\underline{0.7225}} &0.4361  \\
T-RR+MBR w/ COMET &34.20 &63.35 &0.7526 &\textbf{\underline{0.6418}} &29.46 &57.13 &0.7058 &\textbf{\underline{0.5005}} \\
\bottomrule
\end{tabular}
\caption{\label{tab:ranking-and-metrics}Evaluation metrics for $\text{EN}\rightarrow \text{DE}$ for the \textit{large} and \textit{small} model settings, using a \textit{fixed} $N$-best reranker (F-RR), a \textit{tuned} $N$-best reranker (T-RR), MBR decoding, and a two-stage approach. Best overall values are \textbf{bolded} and best for each specific group are \underline{underlined}.} 
\vspace{-0.3cm}
\end{table*}

\begin{figure}[t]
     \centering
      \includegraphics[width=0.95\columnwidth]{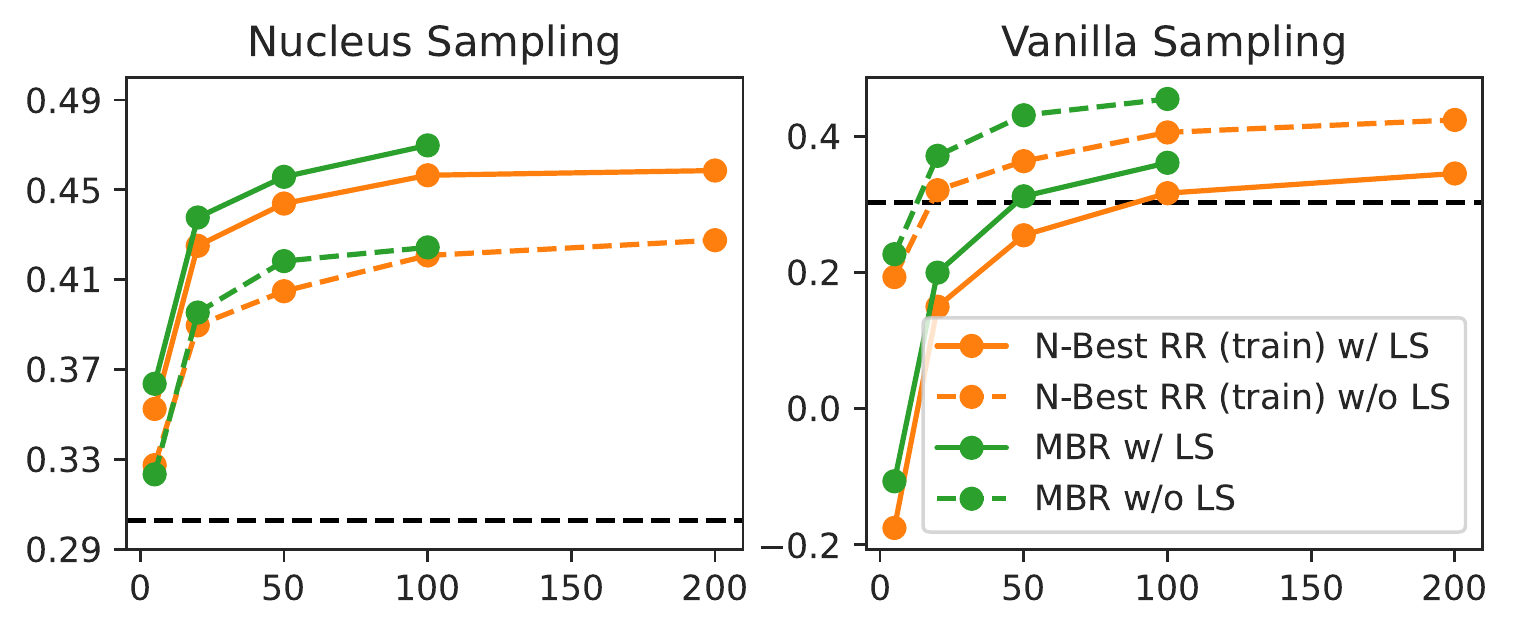}
     \vspace{-0.2em}
     \caption{\label{fig:no-ls-comparison} COMET scores for $\text{EN}\rightarrow \text{DE}$ (IWSLT17) for models trained with and without label smoothing.}
     \vspace{-1em}
\end{figure}
\subsubsection{Impact of Label Smoothing}
\label{subsubsec:label-smoothing}
\paragraph{\textit{How does label smoothing affect candidate generation?}}
Label smoothing \citep{szegedy2016smoothing} is a regularization technique that redistributes probability mass from the gold label to the other target labels, typically preventing the model from becoming overconfident \citep{muller2019when}.
However, it has been found that label smoothing negatively impacts model fit, compromising the performance of MBR decoding \citep{eikema-aziz-2020-map, eikema2021samplingbased}. 
Thus, we train a small transformer model without label smoothing to verify its impact in the performance of $N$-best reranking and MBR decoding.
Figure~\ref{fig:no-ls-comparison} shows that disabling label smoothing really helps when generating candidates using vanilla sampling. However, the performance \emph{degrades} for candidates generated using nucleus sampling when we disable label smoothing, hinting that the pruning mechanism of nucleus sampling may help mitigate the negative impact of label smoothing in sampling based approaches.
Even without label smoothing, vanilla sampling is not competitive with nucleus sampling or beam search with label smoothing, thus, we do not experiment further with it.

\subsubsection{Impact of Ranking and Metrics}

We now investigate the usefulness of the metrics presented in \S\ref{sec:quality-aware} as features and objectives for ranking.
For $N$-best reranking, we use all the available candidates (200) 
while, for MBR, due to the computational cost of using 100 candidates, we report results with 50 candidates only (we found that ranking with \textit{tuned} $N$-best reranking with $N=100$ and MBR with $N=50$ takes about the same time).
We report results in Table~\ref{tab:ranking-and-metrics}, and use them to answer some specific research questions.

\paragraph{\textit{Which QE metric works best in a fixed $N$-best reranker?}}
We consider a \textit{fixed} $N$-best reranker with a single reference-free metric as a feature (see Table~\ref{tab:ranking-and-metrics}, second group).
While none of the metrics allows for improving the baseline results in terms of the lexical metrics (BLEU and chrF), rerankers using COMET-QE or MBART-QE outperform the baseline according to BLEURT and COMET, for both the \emph{large} and \emph{small} models.
Due to the aforementioned better performance of these metrics for translation quality evaluation, we hypothesize that these rankers produce better translations than the baseline. However, since the sharp drop in the lexical metrics is concerning, we will verify this hypothesis in a human study, in \S\ref{subsubsec:human-evaluation}.

\paragraph{\textit{How does the performance of a tuned $N$-best reranker vary when we change the optimization objective?}}
We consider a \textit{tuned} $N$-best reranker using as features \textit{all} the reference-free metrics  in \S\ref{subsec:Reference-free Metrics}, and optimized 
using MERT. Table \ref{tab:ranking-and-metrics} (3\textsuperscript{rd} group) shows  results for $\text{EN}\rightarrow \text{DE}$. 
For the \textit{small} model, all the rankers show improved results over the baseline for all the metrics. In particular, optimizing for BLEU leads to the best results in the lexical metrics, while optimizing for BLEURT leads to the best performance in the others.
Finally, optimizing for COMET leads to similar performance than optimizing for BLEURT. 
For the \emph{large} model, although none of the rerankers is able to outperform the baseline in the lexical metrics, we see similar trends as before for BLEURT and COMET.

\paragraph{\textit{How does the performance of MBR decoding vary when we change the utility function?}}
Table~\ref{tab:ranking-and-metrics} (4\textsuperscript{th} group) shows the impact of the utility function (BLEU, BLEURT, or COMET). For the \emph{small} model, using COMET leads to the best performance according to all the metrics except BLEURT (for which the best result is attained when optimizing itself).
For the \emph{large} model, the best result according to a given metric is obtained when using that metric as the utility function.

\paragraph{\textit{How do (tuned) $N$-best reranking and MBR compare to each other?}}
Looking at Table~\ref{tab:ranking-and-metrics} we see that, for the \emph{small} model, $N$-best reranking seems to perform better than MBR decoding in all the evaluation metrics, including the one used as the utility function in MBR decoding. 
The picture is less clear for the \emph{large} model, with MBR decoding achieving best values for a given fine-tuned metric when using it as the utility; this comes at the cost of worse performance according to the other metrics, hinting at a potential ``\emph{overfitting}'' effect. Overall, $N$-best reranking seems to have an edge over MBR decoding. We will further clarify this question with human evaluation in \S~\ref{subsubsec:human-evaluation}.

\paragraph{\textit{Can we improve performance by combining $N$-best reranking with MBR decoding?}}
Table~\ref{tab:ranking-and-metrics} shows that, for both the \emph{large} and the \emph{small} model, the two-stage ranking approach described in \S\ref{sec:quality-aware} leads to the best performance according to the fine-tuned metrics. In particular, the best result is obtained when the utility function is the same as the evaluation metric.
These results suggest that a promising research direction is to seek more sophisticated pruning strategies for MBR decoding. 

\begin{table*}[t]
\small
\centering
\resizebox{0.95\linewidth}{!}{%
\setlength\tabcolsep{4pt}
\begin{tabular}{lcccccrccccc}\toprule
&\multicolumn{5}{c}{EN-DE (WMT20)} &\multicolumn{5}{c}{EN-RU (WMT20)} \\
\cmidrule(lr){2-6} \cmidrule(lr){7-11}
&BLEU &chrF & BLEURT &COMET &Human R. &BLEU &chrF &BLEURT &COMET &Human R. \\\midrule
Reference &- &- &- &- &4.51 &- &- &- &- &  4.07 \\
\midrule
Baseline &\textbf{36.01} &\textbf{63.88} &0.7376 &0.5795 &4.28 &\textbf{23.86} &51.16 &0.6953 &0.5361 & 3.62\\
F-RR w/ COMET-QE &29.83 &59.91 &0.7457 &0.6012 &4.19 &20.32 &49.18 &0.7130 &0.6207 & 3.25\\
T-RR w/ COMET &34.26 &63.31 &\textbf{0.7546} &0.6276 &\textbf{4.33} &22.42 &50.91 &\textbf{0.7243} &0.6441 & 3.65\\
MBR w/ COMET &33.04 &62.65 &0.7477 &0.6359 &4.27 &23.67 &51.18 &0.7093 &0.6242 & 3.66\\
T-RR + MBR w/ COMET &34.20 &63.35 &0.7526 &\textbf{0.6418} &4.30 &23.21 &\textbf{51.26} &0.7238 &\textbf{0.6736} & \textbf{3.72}$^\dagger$\\
\toprule
&\multicolumn{5}{c}{EN-DE (IWSLT17)} &\multicolumn{5}{c}{EN-FR (IWSLT17)} \\
\cmidrule(lr){2-6} \cmidrule(lr){7-11}
Reference &- &- &- &- &4.38 &- &- &- &- & 4.00\\
\midrule
Baseline &29.12 &0.6635 &56.23 &0.3028 &3.68 &38.12\, &0.6532 &63.20 &0.4809 & 3.92 \\
F-RR w/ COMET-QE &27.38 &0.6848 &54.89 &0.4071 &3.67\, &35.59 &0.6628 &60.90 &0.5553 & 3.63\\
T-RR w/ COMET &\textbf{30.16} &\textbf{0.7124} &\textbf{57.32} &0.4721 &\textbf{3.90}$^\dagger$ &\textbf{38.60} &\textbf{0.7020} &\textbf{63.77} &0.6392 & 4.05$^\dagger$\\
MBR w/ COMET&29.43 &0.6882 &56.74 &0.4480 &3.79$^\dagger$ &37.77 &0.6710 &63.24 &0.6127 & 4.05$^\dagger$\\
T-RR + MBR w/ COMET &29.46 &0.7058 &57.13 &\textbf{0.5005} &3.83$^\dagger$ &38.33 &0.6883 &63.53 &\textbf{0.6610} & \textbf{4.09}$^\dagger$\\
\bottomrule
\end{tabular}
}
\caption{\label{table:results_human_eval} Results for automatic and human evaluation. 
Top: WMT20 (large models); Bottom: IWSLT17 (small models).
Methods with $^\dagger$ are statistically significantly better than the baseline, with $p<0.05$.}
\vspace{-0.4cm}
\end{table*}

\subsubsection{Human Evaluation}
\label{subsubsec:human-evaluation}

\paragraph{\textit{Which metric correlates more with human judgments? How risky is it to optimize a metric and evaluate on a related metric?}}
Our experiments suggest that, overall,
\textit{quality-aware} decoding produces translations with better performance across most metrics than \textit{MAP-based} decoding. However, for some cases (such as fixed $N$-best reranking and most results with the \textit{large} model), there is a concerning ``metric gap'' between lexical-based and fine-tuned metrics. While the latter have shown to correlate better with human judgments, previous work has not attempted to explicitly optimize these metrics, and doing so could lead to ranking systems that learn to exploit ``pathologies'' in these metrics rather than improving translation quality. 
To investigate this hypothesis, we perform a human study across all four datasets. We ask annotators to rate, from 1 (no overlap in meaning) to 5 (perfect translation), the translations produced by the 4 \textit{ranking} systems in \S\ref{sec:quality-aware}, as well as the baseline translation and the reference. Further details are in App.~\ref{sec:appendix_human_study}. 
We choose COMET-QE as the feature for the fixed $N$-best ranker and COMET as the optimization metric and utility function for the tuned $N$-best reranker and MBR decoding, respectively. The reasons for this are two-fold: (1) they are currently the reference-free and reference-based metrics with highest reported correlation with human judgments \citep{kocmi2021ship}, (2) we saw the largest ``metric gap'' for systems based on these metrics, hinting of a potential ``overfitting'' problem (specially since COMET-QE and COMET are  similar models).


Table~\ref{table:results_human_eval} shows the results for the human evaluation, as well as the automatic metrics. We see that, with the exception of T-RR w/ COMET, when fine-tuned metrics are explicitly optimized for, their correlation with human judgments decreases and they are no longer reliable indicators of system-level ranking. This is notable for the fixed $N$-best reranker with COMET-QE, which outperforms the baseline in COMET for every single scenario, but leads to markedly lower quality translations. However, despite the potential for overfitting these metrics, we find that \textit{tuned} $N$-best reranking, MBR, and their combination consistently achieve better translation quality than the baseline, specially with the small model. In particular, $N$-best reranking results in better translations than MBR, and their combination is the best system in 2 of 4 LPs. 

\subsubsection{Improved Human Evaluation}
To further investigate how \emph{quality-aware} decoding performs when compared to \emph{MAP-based} decoding, we perform another human study, this time based on expert-level multidimensional quality metrics (MQM) annotations \citep{mqm}. 
We asked the annotators to identify all errors and independently label them with an error category (\emph{accuracy}, \emph{fluency}, and \emph{style}, each with a specific set of subcategories) and a severity level (\emph{minor}, \emph{major}, and \emph{critical}). In order to obtain the final sentence-level scores, we require a weighting scheme on error severities. We use weights of $1$, $5$, and $10$ to \emph{minor}, \emph{major}, and \emph{critical} errors, respectively, independently of the error category. Further details are in App.~\ref{sec:appendix_human_study_mqm}.
Given the cost of performing a human study like this, we restrict our analysis to the translations generated by the large models trained on WMT20 (EN $\rightarrow$ DE and EN $\rightarrow$ RU).

Table~\ref{table:results_human_eval_additional} shows the results for the human evaluation using MQM annotations, including both error severity counts and final MQM scores.
As hinted in \S\ref{subsubsec:human-evaluation}, despite the remarkable performance of the F-RR with COMET-QE in terms of COMET (see Table~\ref{table:results_human_eval}), the quality of the translations decreases when compared to the baseline, suggesting the possibility of \emph{metric overfitting} when evaluating systems using a single automatic metric that was directly optimized for (or a similar one).
However, for both language pairs, the T-RR with COMET and the two stage approach (T-RR + MBR with COMET) achieve the highest MQM score. In addition, these systems present the smallest number of errors when combining both major and critical errors.

Although the performance of all systems is comparable for EN$\rightarrow$DE, both the T-RR and the T-RR+MBR decoding markedly reduce the number of grammatical register errors related to using pronouns and verb forms that are not compliant with the register required for that translation, at the cost of increasing the number of lexical selection errors (see Figure~\ref{fig:mqm-ende}). For EN$\rightarrow$RU, however, the number of lexical selection errors produced when using the T-RR or the T-RR+MBR decoding is approximately a half of the ones produced by the baseline (see Figure~\ref{fig:mqm-enru}). In this case, this comes at apparently almost no cost in other error types, leading to significantly better results, as shown in Table~\ref{table:results_human_eval_additional}.

\begin{table*}[t]
\small
\centering
\resizebox{0.75\linewidth}{!}{%
\setlength\tabcolsep{4pt}
\begin{tabular}{lccccrcccc}\toprule
&\multicolumn{4}{c}{EN-DE (WMT20)} &\multicolumn{4}{c}{EN-RU (WMT20)} \\
\cmidrule(lr){2-5} \cmidrule(lr){6-9}
&Minor &Major &Critical &MQM &Minor &Major &Critical &MQM \\\midrule
Reference &24 &67 &0 &97.04 &5 &11 &0 &99.30\\
\midrule
Baseline &8 &139 &0 &95.66 &17 &239 &49 &79.78\\
F-RR w/ COMET-QE &15 &204 &0 &93.47 &13 &254 &80 &76.25\\
T-RR w/ COMET &12 &109 &0 &\textbf{96.20} &9 &141 &45 &{85.97}$^\dagger$\\
MBR w/ COMET &11 &161 &0 &94.38 &8 &182 &40 &83.65\\
T-RR + MBR w/ COMET &10 &138 &0 &{95.44} &11 &134 &45 &\textbf{86.78}$^\dagger$\\
\bottomrule
\end{tabular}}
\caption{\label{table:results_human_eval_additional} Error severity counts and MQM scores for WMT20 (large models). Best overall values are \textbf{bolded}. Methods with $^\dagger$ are statistically significantly better than the baseline, with $p<0.05$.}
\vspace{-0.4cm}
\end{table*}

\section{Related Work}
\label{sec:related}

\paragraph{Reranking.}
Inspired by the work of \citet{shen-etal-2004-discriminative} on discriminative reranking for SMT, \citet{lee-etal-2021-discriminative} trained a large transformer model using a reranking objective to optimize BLEU. Our work differs in which our rerankers are much simpler  and therefore can be tuned on a validation set; and we use more powerful quality metrics instead of BLEU. Similarly, \citet{bhattacharyya-etal-2021-energy} learned an energy-based reranker to assign lower energy to the samples with higher BLEU scores.
While the energy model plays a similar role to a QE system, our work differs in two ways: we use an existing, pretrained QE model instead of training a dedicated reranker, making our approach applicable to any MT system without  further training; and the QE model is trained to predict human assessments, rather than BLEU scores.  
\citet{leblond-etal-2021-machine} compare a reinforcement learning approach to  reranking approaches (but not MBR decoding, as we do). They investigate the use of reference-based metrics and, for the reward function, a reference-free metric based on a modified BERTScore~\cite{Zhang2020BERTScore}. 
This new multilingual BERTScore is not fine-tuned on human judgments as COMET and BLEURT and it is unclear what its level of agreement with human judgments is. 
Another line of work is \textit{generative reranking}, where the reranker is not trained to optimize a  metric, but rather as a generative noisy-channel model \citep{yu-etal-2017-noisy, yee-etal-2019-simple, ng-etal-2019-facebook}. 

\paragraph{Minimum Bayes Risk Decoding.}
MBR decoding \citep{kumar2002minimum, kumar-byrne-2004-minimum} has recently been revived for NMT using candidates generated with beam search \citep{stahlberg-etal-2017-neural, shu2017laterstage} and sampling \citep{eikema-aziz-2020-map, muller2021understanding}. \citet{eikema2021samplingbased} also explore a two-stage approach for MBR decoding.
Additionally, there is concurrent work by \citet{freitag2021minimum} on using neural metrics as utility functions during MBR decoding: however they limit their scope to MBR with reference-based metrics, while we perform a more extensive evaluation over ranking methods and metrics. \citet{chantal2022weakness} also concurrently explored using MBR decoding with neural metrics, but with the purposes of identifying weaknesses in the metric (in their case COMET), similarly to the \textit{metric overfitting} problem we discussed in \S\ref{subsubsec:human-evaluation}.
A comparison 
with $N$-best  re-ranking was missing in these  works, a gap our paper fills.
A related line of work is \textit{minimum risk training} (MRT; \citealt{smith-eisner-2006-minimum,shen-etal-2016-minimum}), which \textit{trains} models to minimize risk, allowing arbitrary non-differentiable loss functions  \citep{edunov-etal-2018-classical, wieting-etal-2019-beyond} and avoiding exposure bias \citep{wang-sennrich-2020-exposure, kiegeland-kreutzer-2021-revisiting}. However, MRT is considerably more expensive and difficult to train and the gains are often small. 
Incorporating our quality metrics in MRT is an exciting research direction.


\section{Conclusions and Future Work}
\label{sec:conclusions}
We leverage recent advances in MT quality estimation and evaluation and propose \emph{quality-aware decoding} for NMT.
We explore different candidate generation and ranking methods, with a comprehensive empirical analysis across four datasets and two model classes.
We show that, compared to MAP-based decoding, quality-aware decoding leads to better translations, according to powerful automatic evaluation metrics and human judgments.

There are several directions for future work. Our ranking strategies increase accuracy but are substantially more expensive, particularly when used with costly  metrics such as BLEURT and COMET. While reranking-based pruning before MBR decoding was found helpful, additional strategies such as caching encoder representations~\cite{chantal2022weakness} and distillation  \citep{pu-etal-2021-learning} are promising directions.


\section*{Acknowledgments}
We would like to thank Ben Peters, Wilker Aziz, and the anonymous reviewers for useful feedback. 
This work was supported by the P2020 program MAIA (LISBOA-01-0247-
FEDER-045909), the European Research Council (ERC StG DeepSPIN 758969), the European Union's Horizon 2020 research and innovation program (QUARTZ grant agreement 951847), and by the Fundação para a Ciência e Tecnologia through UIDB/50008/2020.

\bibliography{anthology,custom}
\bibliographystyle{acl_natbib}

\appendix

\onecolumn

\clearpage

\newpage

\noindent {\Large{\textbf{Supplemental Material}}}

\section{Training Details}
\label{sec:training_details}

For the experiments using IWSLT17, we train a \emph{small} transformer model (6 layers, 4 attention heads, 512 embedding dimensions, and 1024 hidden dimensions) from scratch, using \textit{Fairseq} \citep{ott-etal-2019-fairseq}. We tokenize the data using SentencePiece \citep{kudo-richardson-2018-sentencepiece}, with a joint vocabulary with 20000 units. We train using the Adam optimizer \citep{kingma2015adam} with $\beta_1=0.9$ and $\beta_2=0.98$ and use an inverse square root learning rate scheduler, with an initial learning rate of $5\times 10^{-4}$ and with a linear warm-up in the first $4000$ steps. For models trained with label smoothing, we use the default value of $0.1$.

\section{Additional Results}
\label{sec:appendix_full_results}
For completeness, we include in Table~\ref{tab:ranking-and-metrics-app} results to evaluate the impact of the metrics presented in \S\ref{sec:quality-aware} as features and objectives for ranking using the other language pairs: $\text{EN}\rightarrow \text{RU}$ (large model) and $\text{EN}\rightarrow \text{FR}$ (small model).

\begin{table*}[h]\centering
\small
\begin{tabular}{lrrrrrrrrr}\toprule
&\multicolumn{4}{c}{Large (WMT20)} &\multicolumn{4}{c}{Small (IWSLT)} \\\cmidrule(lr){2-5} \cmidrule(lr){6-9}
&BLEU &chrF &BLEURT &COMET &BLEU &chrF &BLEURT &COMET \\\midrule
Baseline &23.86 &51.16 &0.6953 &0.5361 &38.12 &63.20 &0.6532 &0.4809 \\
\midrule
F-RR w/ COMET-QE &20.32 &49.18 &\underline{0.7130} &\underline{0.6207} &35.59 &60.90 &0.6628 &\underline{0.5553} \\
F-RR w/ MBART-QE &\underline{22.39} &\underline{50.59} &0.6993 &0.5481 &\underline{36.68} &\underline{62.17} &0.6593 &0.5091 \\
F-RR w/ OpenKiwi &20.88 &48.72 &0.7040 &0.5688 &32.03 &55.68 &0.5996 &0.2581 \\
F-RR w/ Transquest &21.60 &50.14 &0.7060 &0.5836 &36.02 &62.26 &\underline{0.6681} &0.5397 \\
\midrule
T-RR w/ BLEU &\underline{23.87} &\underline{\textbf{51.51}} &0.7042 &0.5669 &\underline{\textbf{39.10}} &\underline{\textbf{64.22}} &0.6968 &0.6189 \\
T-RR w /BLEURT &22.84 &51.25 &\underline{0.7265} &\underline{0.6470} &38.60 &63.76 &\underline{0.7042} &\underline{0.6405} \\
F-RR w/ COMET &22.42 &50.91 &0.7243 &0.6441 &38.60 &63.77 &0.7020 &0.6392 \\
\midrule
MBR w/ BLEU &\underline{24.03} &51.12 &0.6938 &0.5393 &\underline{37.97} &63.13 &0.6484 &0.4764 \\
MBR w/ BLEURT &23.01 &50.87 &\underline{0.7314} &0.5984 &37.29 &62.82 &\underline{0.6886} &0.5361 \\
MBR w/ COMET &23.67 &\underline{51.18} &0.7093 &\underline{0.6242} &37.77 &\underline{63.24} &0.6710 &\underline{0.6127} \\
\midrule
T-RR+MBR w/ BLEU &\underline{\textbf{24.11}} &\underline{51.44} &0.6967 &0.5482 &\underline{38.96} &\underline{64.04} &0.6781 &0.5636 \\
T-RR+MBR w/ BLEURT &23.18 &51.30 &\underline{\textbf{0.7344}} &0.6277 &37.43 &63.14 &\underline{\textbf{0.7092}} &0.5961 \\
T-RR+MBR w/ COMET &23.21 &51.26 &0.7238 &\underline{\textbf{0.6736}} &38.33 &63.53 &0.6883 &\underline{\textbf{0.6610}} \\
\bottomrule
\end{tabular}
\caption{\label{tab:ranking-and-metrics-app}Evaluation metrics for $\text{EN}\rightarrow \text{RU}$ for the \textit{large} model setting and  $\text{EN}\rightarrow \text{FR}$ for \textit{small} model settings, using a \textit{fixed} $N$-best reranker (F-RR), a \textit{tuned} $N$-best reranker (T-RR), MBR decoding, and a two-stage approach. Best overall values are \textbf{bolded} and best for each specific group are \underline{underlined}.} 
\end{table*}




\section{Human Study}
\label{sec:appendix_human_study}

In order to perform human evaluation, we recruited professional translators who were native speakers of the target language on the freelancing site Upwork.\footnote{\url{https://upwork.com}. Freelancers were paid a market rate of 18-20 US dollars per hour, and finished approximately 50 sentences in one hour.}
300 sentences were evaluated for each language pair, sampled randomly from the test sets after a restriction that sentences were no longer than 30 words.
All translation hypotheses for a single source sentence were first deduplicated, and then shown to the translator side-by-side in randomized order to avoid any ordering biases.

Sentences were evaluated according to a 1-5 rubric slightly adapted from that of \citet{wieting-etal-2019-beyond}:
\begin{enumerate}
    \item There is no overlap in the meaning of the source sentence whatsoever.
    \item Some content is similar but the most important information in the sentence is different.
    \item The key information in the sentence is the same but the details differ.
    \item Meaning is essentially equal but some expressions are unnatural.
    \item Meaning is essentially equal and the sentence is natural.
\end{enumerate}

\section{MQM Framework}
\label{sec:appendix_human_study_mqm}

Human evaluations were performed by Unbabel's PRO Community, made of professional translators and linguists with relevant experience in linguistic annotations and translation errors annotations. In order to properly assess translations quality, annotators must be native speakers of the target language and with a proven high proficiency of the source language, so that they can properly capture errors and their nuances. The systems' outputs were evaluated by using the annotation framework adopted internally at Unbabel, which is an adaptation of the MQM Framework \citep{mqm}.

We asked the annotators to identify all errors and independently label them with an error category and a severity level. We consider \textbf{three categories} (each of them containing a set of different subcategories) that may affect the quality of the translations:
\begin{itemize}
    \item \emph{Accuracy}, if the target text does not accurately reflect the source text (\textit{e.g.}, changes in the meaning, addition/omission of information, untranslated text, MT hallucinations);
    \item \emph{Fluency}, if there are issues that affect the reading and the comprehension of the text (\textit{e.g.}, grammar and spelling errors); 
    \item \emph{Style}, if the text has stylistic problems (\textit{e.g.}, gramatical and lexical register).
\end{itemize}
Additionally, each error is labeled according to \textbf{three severity levels} (\emph{minor}, \emph{major}, and \emph{critical}), depending on the way they affect the accuracy, the fluency, and the style of the translation. The final sentence-level score is obtained using a weighting scheme where minor, major, and critical errors are weighted as $1$, $5$, and $10$, respectively.

\begin{figure}
    \centering
    \includegraphics[width=\textwidth]{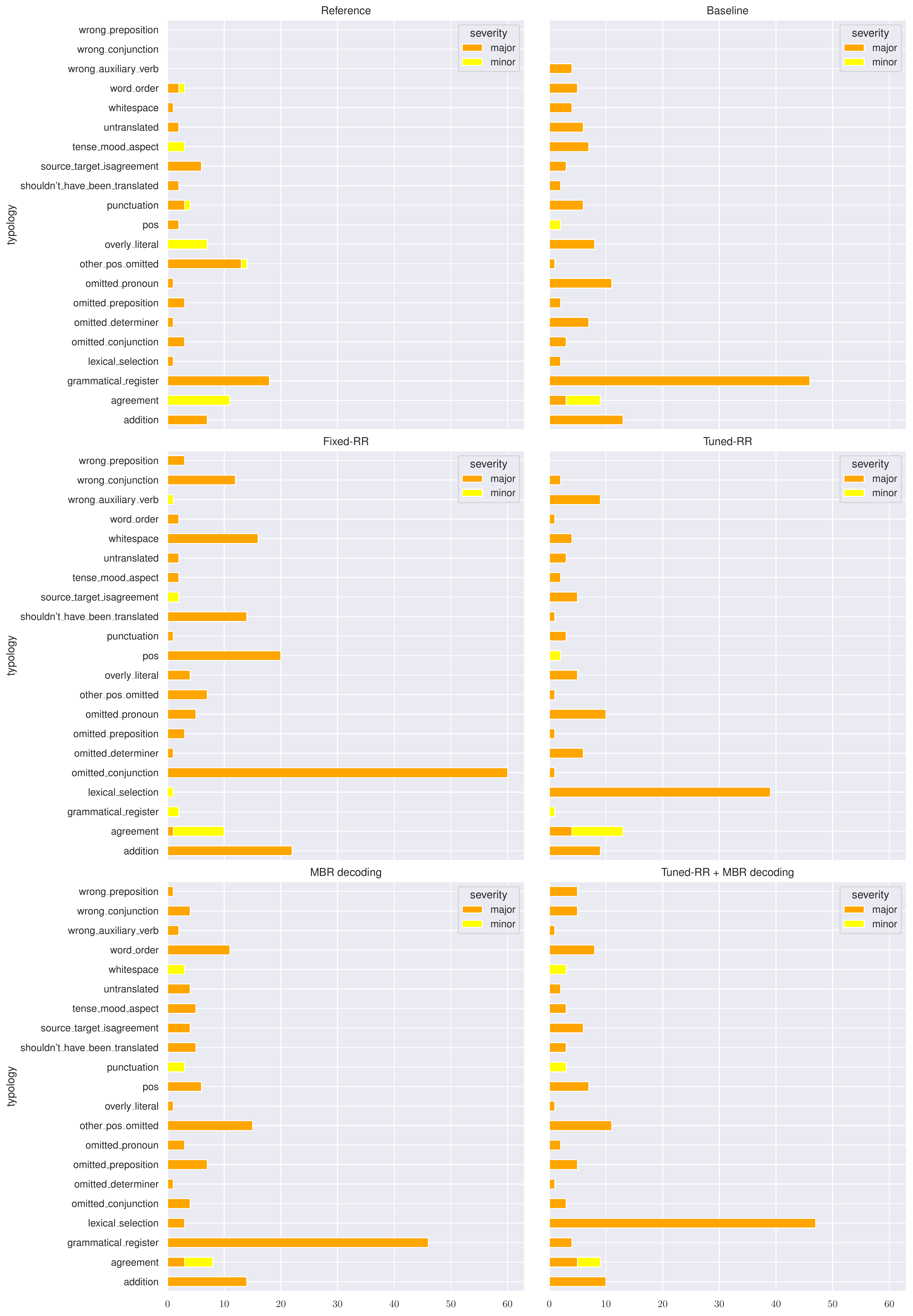}
    \caption{Error typology and severity level breakdown for WMT20 (large models) EN$\rightarrow$DE.}
    \label{fig:mqm-ende}
\end{figure}

\begin{figure}
    \centering
    \includegraphics[width=\textwidth]{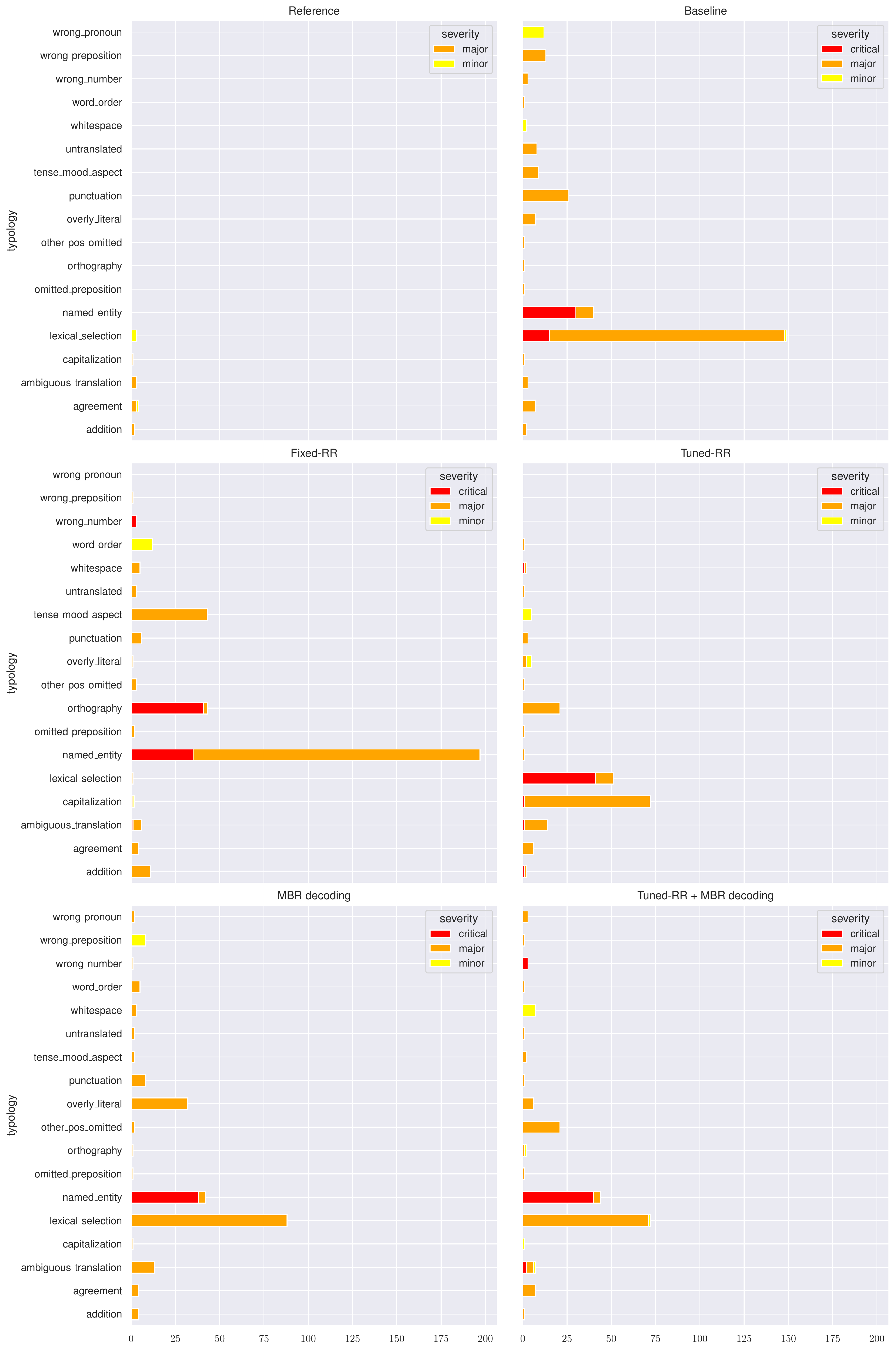}
    \caption{Error typology and severity level breakdown for WMT20 (large models) EN$\rightarrow$RU.}
    \label{fig:mqm-enru}
\end{figure}

Figures~\ref{fig:mqm-ende} and~\ref{fig:mqm-enru} show the counts of errors breakdown by typology and severity level for EN$\rightarrow$DE and EN$\rightarrow$RU, respectively.

\end{document}